\documentclass[letterpaper, 10 pt, journal, twoside]{ieeetran} 

\IEEEoverridecommandlockouts

\usepackage{etextools}
\usepackage{amssymb}
\usepackage{float}
\usepackage{makeidx}
\usepackage{amsmath}
\usepackage{bbm}
\usepackage{epsfig}
\usepackage{epsf}
\usepackage{psfrag}
\usepackage{verbatim}
\usepackage{color}
\usepackage{multirow}
\usepackage{tabularx}
\usepackage{booktabs}
\usepackage[tight,footnotesize]{subfigure}
\usepackage{array}
\usepackage{soul}
\usepackage{footnote}
\usepackage{cite}
\usepackage{dblfloatfix}
\usepackage{xcolor}
\usepackage{color, colortbl}
\usepackage[colorlinks,bookmarksnumbered,citecolor=orange,urlcolor=orange]{hyperref}
\usepackage{graphicx}
\graphicspath{{Figures}}
\DeclareGraphicsExtensions{.pdf,.png}
\usepackage{bigstrut}
\usepackage[english]{babel}
\usepackage{csquotes}
\usepackage{balance}
\usepackage[T1]{fontenc}
\usepackage{algorithm, setspace}
\usepackage{algpseudocode}
\usepackage{url}
\usepackage{multirow}
\usepackage{float}
\usepackage{xcolor}
\usepackage{hyperref}
 \hypersetup{
     colorlinks=true,
     linkcolor=orange,
     filecolor=orange,
     citecolor=orange,      
     urlcolor=orange,
     }

\newcommand{\p}[1]{\smallskip \noindent \textbf{{#1}.}}
\newcommand{\eq}[1]{Equation~(\ref{eq:#1})}
\newcommand{\fig}[1]{Fig.~\ref{fig:#1}}

\title{

Communicating Inferred Goals with Passive Augmented Reality and Active Haptic Feedback

}

\author{James F. Mullen Jr, Josh Mosier, Sounak Chakrabarti, Anqi Chen, Tyler White, and Dylan P. Losey
\thanks{The student authors were equal members of a senior design team.}
\thanks{The authors are with the Collaborative Robotics Lab,
Department of Mechanical Engineering, Virginia Tech, Blacksburg, VA 24060 USA (e-mail: {\tt\footnotesize mullenj@vt.edu}, {\tt\footnotesize losey@vt.edu}).}
}

\begin{document}
\maketitle

\begin{abstract}

Robots learn as they interact with humans. Consider a human teleoperating an assistive robot arm: as the human guides and corrects the arm’s motion, the robot gathers information about the human’s desired task. But how does the \textit{human} know what their robot has inferred? Today’s approaches often focus on conveying intent: for instance, using legible motions or gestures to indicate what the robot is planning. However, closing the loop on robot inference requires more than just revealing the robot’s current policy: the robot should also display the alternatives it thinks are likely, and prompt the human teacher when additional guidance is necessary. In this paper we propose a multimodal approach for communicating robot inference that combines both passive and active feedback. Specifically, we leverage information-rich augmented reality to \textit{passively visualize} what the robot has inferred, and attention-grabbing haptic wristbands to \textit{actively prompt} and direct the human’s teaching. We apply our system to shared autonomy tasks where the robot must infer the human’s goal in real-time. Within this context, we integrate passive and active modalities into a single algorithmic framework that determines \textit{when} and \textit{which type} of feedback to provide. Combining both passive and active feedback experimentally outperforms single modality baselines; during an in-person user study, we demonstrate that our integrated approach increases how efficiently humans teach the robot while simultaneously decreasing the amount of time humans spend interacting with the robot. Videos here: \url{https://youtu.be/swq_u4iIP-g}

\end{abstract}


\smallskip


\section{Introduction}

\begin{figure}[t]
	\begin{center}
		\includegraphics[width=0.8\columnwidth]{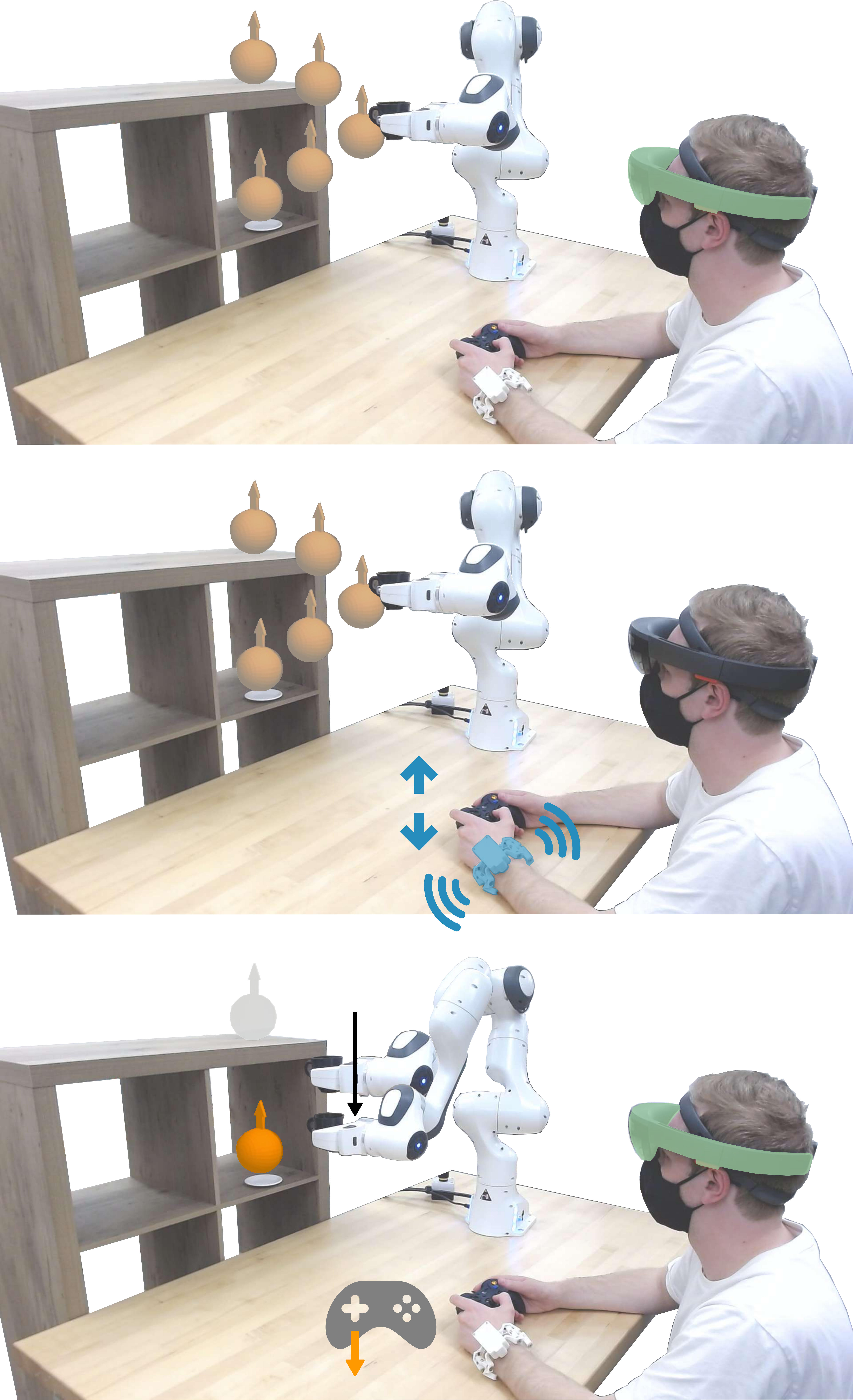}
		\caption{Human teleoperating a $7$-DoF robot arm. (Top) The robot believes it should place the cup on a shelf, and passively visualizes the likely options in augmented reality. (Middle) But the robot is still uncertain about the right shelf. To elicit informative feedback, the robot actively signals the haptic wristband, and prompts the human to guide the robot either up or down. (Bottom) The human responds by moving down. In augmented reality, the robot shows that it has inferred the bottom shelf is the human's goal.}
		\label{fig:front}
	\end{center}
	\vspace{-2.25em}
\end{figure}

\IEEEPARstart{I}{magine} teaching an assistive robot arm how to put away a coffee cup (see \fig{front}). You want the robot to place this cup upright on the lower shelf --- as the robot moves toward the shelves, you correct its mistakes and guide the robot towards your desired goal. The robot gathers information in real-time from your corrections, updating its understanding of how and where it should carry the coffee cup. But how do \textit{you} know what the robot has \textit{inferred}? Ideally you would like to know when you have taught the robot enough, so that you can step away and do something else. But without feedback you are stuck watching the robot throughout the entire task, unsure whether it fully understands your goal, or what corrections you should provide to remove any remaining confusion!

Existing research enables robots to convey their \textit{intent} to nearby humans \cite{hellstrom2018understandable,dragan2013legibility,tellex2014asking,admoni2016predicting,cha2018survey}. These prior works study a spectrum of different modalities, including legible motions, language, gaze, gestures, and displays. Each new modality offers a separate channel for providing intent feedback to the human. But communicating inferred behavior requires more than just intent: the robot learner must indicate what it thinks it should do, display its confidence that this inferred behavior is correct, and even prompt additional human teaching when unsure. As robot learners move from communicating their immediate intent to proactively involving human teachers, we propose that robots synthesize \textit{multiple} modalities to close-the-loop and make it clear what they do and do not know.

Selecting the right modalities often depends on the current user and task. But --- whichever modalities we choose --- our central hypothesis is that combining passive and active feedback will better bring the human into the learning loop than systems which only leverage a single feedback type:
\begin{center}\vspace{-0.3em}
\textit{Robots should \emph{passively} convey what they have inferred, and \emph{actively} prompt the human when teaching is required.}
\vspace{-0.3em}
\end{center}
Specifically, in this paper we integrate augmented reality displays and wearable haptic feedback to develop a multimodal interface that passively communicates learning and actively elicits teaching. We apply our system to shared autonomy settings \cite{jain2019probabilistic, javdani2018shared, dragan2013policy}. Here the robot arm knows a discrete set of possible goals, and infers the desired goal from the human's teleoperation inputs. As shown in \fig{front}, our interface \textit{passively displays} the different ways to complete the task that the robot thinks are likely, and \textit{actively prompts} the human when further teaching is required. We conduct user studies to test how this feedback system affects interactions between human teachers and robot learners. When compared to single modality baselines, we find that the combination of passive and active feedback i) prevents users from continuing to teach the robot after it has already learned what they want, ii) reduces the amount of time users spend monitoring the robot, and iii) improves user teaching so that the robot infers what the human wants from fewer interactions.

Overall, our proposed system is a step towards transparent robot learning and accelerated human teaching. We make the following contributions:

\p{Combining Multiple Feedback Modalities} Unlike related works that only consider a single modality, we integrate an existing augmented reality head-mounted display with a novel haptic wristband to convey multimodal information and proactive alerts about robot inference to human teachers.

\p{Converting Robot Inference to Passive and Active Feedback} Specifically in the context of shared autonomy, we formulate an approach that maps robot inference to intuitive multimodal cues. The resulting algorithm determines when to provide which type of feedback, identifies critical states where additional human teaching is especially important, and prompts the most informative direction of human input.

\p{Conducting a User Study} We compare passive and active feedback to single-modality visual, haptic, and augmented reality baselines across an in-person user study. Participants traded-off between teaching the robot arm and performing a distractor task. We find that our approach reduces the amount of time a user must spend focused on the robot learner, and also improves the user's teaching efficiency.
\section{Related Work}

We build on prior work towards explainable robot learners. This includes research that generally communicates robot intent to nearby humans, research that specifically focuses on augmented reality and haptic displays, and research that reveals what the robot is inferring during shared autonomy.

\p{Communicating Robot Intent} For humans and robots to seamlessly collaborate it must be clear to the human what the robot is trying to do \cite{hellstrom2018understandable}. Recent works have developed a variety of different modalities that robots leverage to convey their intent. For instance, robots can use legible \textit{motions} to make their goals clear to human onlookers \cite{dragan2013legibility, szafir2014communication}, or incorporate natural \textit{language} to communicate their objective \cite{tellex2014asking, knepper2017implicit}. Non-verbal communication such as \textit{gestures} \cite{cha2018survey, gielniak2011generating} or \textit{gaze} \cite{admoni2016predicting} are also expressive channels for indicating where the robot plans to go. Besides these implicit cues, robots employ explicit indicators in the form of \textit{projections} \cite{andersen2016projecting, baraka2016enhancing} --- where they render 2D images onto the environment --- or on-board \textit{lights} \cite{weng2019robot} --- which are functionally similar to turn signals on cars.

But while many distinct modalities are available for conveying \textit{intent}, we emphasize that this is different from communicating robot \textit{inference}. Think of our motivating example: communicating the robot's intent is equivalent to showing the final cup location and orientation that the robot thinks is most likely and currently intends to reach. Yet this alone is not enough to make the robot's learning transparent: we may also need to highlight the alternative goals and trajectories the robot is reasoning over, or ask the human for additional feedback to determine which goal is correct.

\p{Augmented Reality and Haptics} In order to convey detailed and attention grabbing signals about robot inference we turn to augmented reality and haptic feedback. Augmented reality head-mounted displays have previously been used to facilitate human-robot interaction \cite{newbury2021visualizing, hoang2021virtual}. Most relevant here are \cite{walker2018communicating, rosen2019communicating, muhammad2019creating, chandan2021arroch}, where the authors leverage augmented reality to convey the robot's internal state, visualize where the robot plans to go, and establish bidirectional communication with the human. This visual overlay succeeds when users are looking directly at the robot --- but when users look away, these graphics are no longer in view. We explore settings where the human must split their attention between the robot and other tasks. Accordingly, we incorporate haptic wristbands that humans wear without encumbering their hands. Recent works have designed similar wristbands to provide tactile feedback for augmented reality \cite{pezent2019tasbi, aggravi2018design}.

Overall, today's research on augmented reality and haptic feedback either develops these two modalities separately, or else utilizes haptics to make interactions with the virtual world seem more realistic. By contrast, we combine the two modalities into a \textit{single} framework to close-the-loop during robot inference. Both active and passive feedback are important because they serve complementary roles: we leverage augmented reality to visualize the high-dimensional behaviors the robot has inferred, and apply haptics to spark and direct user inputs when additional teaching is required.

\p{Revealing Robot Inference} Creating learners that are interpretable by human users is an active area of study \cite{arrieta2020explainable}. For instance, in \cite{huang2019enabling} the robot displays informative examples of its learned behavior on a computer screen, enabling users to extrapolate the robot's underlying objective. Understanding robot learning becomes especially challenging when that learning consists of networks with thousands of weights. However, shared autonomy requires only \textit{low-dimensional} inference: here the robot is inferring the human's desired goal from a discrete set of options \cite{jain2019probabilistic, javdani2018shared, dragan2013policy}. Related work has taken advantage of this structure to visualize the robot's inferred goal using augmented reality \cite{brooks2020visualization, zolotas2018head}.

\begin{figure*}[t]
	\begin{center}
		\includegraphics[width=1.8\columnwidth]{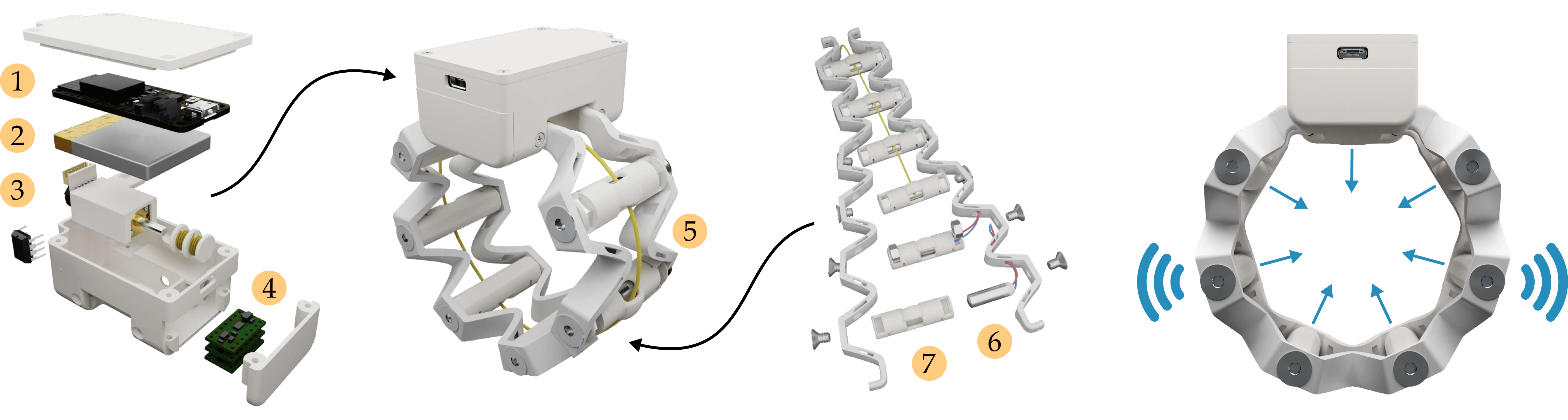}
		\vspace{-1em}
		\caption{Our haptic wristband for delivering active feedback. This device can both gently squeeze the wearer's wrist and render localized vibration patterns. \textbf{Left:} Exploded view of the electronics housing. The microcontroller (1) sits over the battery (2) and squeeze mechanism (3). Three motor drivers responsible for controlling vibration (4) are accessible behind the front panel. \textbf{Middle-Left:} Fully assembled haptic device with cord (5). \textbf{Middle-Right:} Vibrotactors (6) are held in place by printed casings (7). \textbf{Right:} To prompt the human we squeeze their wrist and activate multiple vibrotactors in a pre-defined pattern. For instance, powering the left and right vibrotactors encourages the human to guide the robot either right or left.}
		\label{fig:haptic}
	\end{center}
	\vspace{-2em}

\end{figure*}

Importantly, this related research provides purely \textit{passive} feedback to the user: i.e., while the user watches the robot they see overlays indicating where the robot thinks it should go. But we are also interested in \textit{active} feedback which influences the human to provide new and informative demonstrations. We envision a system that prompts the human when the robot is confused, and even suggests to the human the direction of input that would best reduce the robot's uncertainty about the human's desired goal.

\section{Developing Multimodal Interfaces for Communicating Robot Inference}

We seek transparent robot learners that involve humans in the teaching process. Our hypothesis is that we can better the human into the learning loop with multimodal feedback --- both in the form of passive visualizations, which display the goals that the robot has inferred, and active prompts, which direct the human towards effective teaching inputs. In this section we develop complementary interfaces for each type of feedback. We particularly focus on the design of a custom haptic wristband for eliciting human teaching.

\subsection{Passive Visualization: Augmented Reality}

We start by leveraging an existing augmented reality head-mounted display (Microsoft HoloLens) to \textit{overlay} graphics onto the real-world environment. These visualizations can passively convey complex information --- for example, as the human looks at the robot arm, they see high-dimensional trajectories that the robot has learned from their demonstrations. But how should we design these visualizations to intuitively convey what the robot does and does not know?

In order to create user-friendly visualizations we take inspiration from successful prototypes in prior work \cite{rosen2019communicating, brooks2020visualization, zolotas2018head}. After implementing these prototypes across short pilot tests --- where humans teleoperated the robot arm to perform simple shared autonomy tasks --- we arrived at a design similar to \texttt{NavPoints} \cite{walker2018communicating}. As shown in \fig{front}, here the robot augments the environment by displaying the position and orientation of end-effector waypoints along its likely trajectories. The position of each waypoint is marked with a \textit{sphere}, while its orientation is denoted by an \textit{arrow}. 

This visualization design is suitable for shared autonomy tasks where the robot reasons over multiple possible goals. Returning to our motivating example, imagine the robot is unsure whether it should place the cup on the top shelf or the bottom shelf: our design visualizes this uncertainty through two trajectories starting at the robot's end-effector and going to both of the shelves. The starting point of every trajectory is affixed to the robot's end-effector and moves with that robot. When the robot updates its understanding of the human's goal, we now vary the \textit{color} and \textit{transparency} of each trajectory to passively signal what the robot has inferred: our algorithm for mapping inference to this augmented reality visualization is introduced in Section~\ref{sec:method}.

\subsection{Active Prompts: Haptic Wristband}

To provide haptic feedback we designed and built a low-cost, wireless, and lightweight wristband (see \fig{haptic}). Users wear this device just like a watch. The wristband provides two separate types of active haptic signals: gently \textit{squeezing} the human's wrist and \textit{vibrating} in patterns distributed around the band. Together, these two alerts convey low-dimensional and targeted information. For instance, when the robot needs additional guidance, the device lightly squeezes the human's wrist and prompts the human by vibrating in the direction where user inputs would be most informative.

\p{Squeeze Feedback} The haptic wristband generates squeeze forces through a cord and reel system. This system is controlled by a high torque N$20$ DC motor which sits centrally in the main housing of the device (refer to \fig{haptic}). When we actuate the DC motor in one direction, the cord is retracted into the central housing equally from both sides; actuating the motor in the opposite direction releases the cord and relaxes the band. The cord is routed through the components of the haptic device and is never in direct contact with the wearer's skin. To make this device more comfortable, we normally keep the band in a relaxed state, and only squeeze on the human's wrist at short intervals to provide notifications.

\p{Vibrotactile Feedback} Localized vibrations are output by six cylindrical ERM motors that are evenly spaced around our flexible 3D printed wristband. Each ERM motor (i.e., each vibrotactor) sits rigidly in a resin-printed housing which transfers vibrations directly onto the wearer's skin. We created a compliant accordion-style wristband so that the band fits varying wrist shapes while isolating vibrations from adjacent vibrotactors. Activating multiple vibrotactors in sequence creates patterns which carry additional information: for example, we can activate the top and bottom vibrotactors to indicate to the human to guide the robot either up or down.

\section{Mapping Robot Inference to \\ Passive and Active Feedback} \label{sec:method}

So far we have worked on two feedback modalities in parallel: we created visual designs for existing augmented reality displays and developed a complementary haptic wristband. In this section we \textit{combine} these two feedback mechanisms into a single formulation for communicating robot inference within shared autonomy settings. We derive a decision rule for providing each type of feedback --- augmented reality is used for \textit{passive}, information rich visualizations of goals the robot thinks are likely, while the haptic wristband is used to \textit{actively} signal the user and elicit additional teaching.

\p{Setting} We start by returning to our example from \fig{front}. Here the human is teleoperating a robot arm, and the robot is inferring the human's goal to autonomously assist the human. Let $s \in \mathcal{S}$ be the robot's state (e.g., end-effector position), let $a \in \mathcal{A}$ be the robot's action (e.g., end-effector velocity), and let transition function $T(s, a)$ capture the robot's dynamics. We consider systems with continuous state and action spaces. The human has in mind a goal $g \in \mathcal{G}$ that they want the robot arm to reach (e.g., \textit{placing the cup on the lower shelf}). Their reward function $R_g(s, a)$ depends upon this chosen goal. The robot seeks to maximize the human's cumulative reward subject to uncertainty over the human's goal\footnote{This is an instance of a partially observable Markov decision process where $g$ is the unknown state and Eq.~(\ref{eq:V2}) is the observation model  \cite{javdani2018shared}.}.

Consistent with prior work on shared autonomy \cite{jain2019probabilistic, javdani2018shared, dragan2013policy}, we assume that the set of possible goals $\mathcal{G}$ is \textit{discrete} and the robot knows $\mathcal{G}$ \textit{a priori}. In practice, these goals can be anywhere in the robot's workspace: the only constraint is that the robot must know the possible goal locations $\mathcal{G}$ prior to interaction. But while the robot knows the human's candidate goals, it \textit{does not} know the human's goal during the current task. Instead, the robot maintains a belief $b$ over the set of possible goals. This belief is a probability distribution, i.e., $b(g) = 1$ indicates that the robot is completely convinced that $g \in \mathcal{G}$ is the human's current goal.

Without any insight from the human the robot would have no way of updating $b$ and inferring the preferred goal. But the robot is not performing this task alone --- both the human and robot share control over the robot's motion. The human applies joystick inputs to teleoperate the robot arm, so that $a_h \in \mathcal{A}$ is the human's commanded action. The robot linearly blends this human input with its own autonomous assistance $a_r \in \mathcal{A}$ to get the overall action \cite{jain2019probabilistic, dragan2013policy}:
\begin{equation} \label{eq:SA1}
    a = (1 - \alpha)\cdot a_h + \alpha \cdot a_r
\end{equation}
where $\alpha \in [0,1]$ arbitrates control between human and robot. During our user study we select $\alpha = 0.4$ so that the human's actions are given more weight than the robot's assistance.

If action $a_h$ results from the human's teleoperation input, how do we obtain $a_r$? The robot selects this autonomous action to assist the human based on what it has inferred so far. Assuming that the robot will fully observe the human's desired goal at the next timestep, the optimal autonomous action at state $s$ given belief $b$ becomes \cite{javdani2018shared}:
\begin{equation} \label{eq:SA2}
    a_r = \text{arg}\max_{a \in \mathcal{A}} ~\sum_{g \in \mathcal{G}} b(g) \cdot Q_g(s, a)
\end{equation}
Here $Q_g(s, a)$ is the cumulative reward of taking action $a$ in state $s$ and then optimally completing the task to reach goal $g$. We use $Q_g(s, a) = -\|T(s, a) - g\|^2$ in our user studies to compute this in closed form and perform inference at $1$ kHz. Intuitively, \eq{SA2} encourages the robot to assist across all \textit{likely} goals. As the robot becomes increasingly confident in a particular goal $g$, it leverages \eq{SA2} to move directly towards that goal.

\subsection{Conveying Inference with Passive Visualizations}

The robot infers the human's preferred goal by observing their teleoperation commands $a_h$. We propose to utilize augmented reality head-mounted displays to visualize what the robot is inferring. Augmented reality is suitable here because it offers a user-friendly and intuitive way to convey dense, high-dimensional information (e.g., trajectories). 

\p{Belief Conditioning} Jumping back to our motivating example, imagine that you are trying to teach the robot arm. At the start of the task the robot has a uniform belief $b$ over different ways to place the cup on the shelves. But you do not know this --- perhaps you think you need to teach the robot to move towards the shelves in the first place. To reveal what the robot already knows, here the visualization should show multiple, equally likely trajectories reaching for the shelves. As the robot starts moving to the shelves, you begin to correct its behavior, and the robot updates it's belief to place higher likelihood on your desired goal. Now the visualization should \textit{change}: instead of showing equal trajectories for every goal, the robot should emphasize the trajectories that correspond to your likely goal(s). Put another way, the robot's visualization not only depends on where the robot and goals are located, but it also \textit{depends} on the robot's \textit{belief}.

\p{Inference} In order to display the robot's inference we must \textit{map} beliefs to visualizations. Before we do this, however, we first need to determine how the robot infers the human's goal. Here we apply Bayesian inference to extract the human's goal from their action inputs. Assuming that the human's inputs $a_h$ are conditionally independent, we reach:
\begin{equation} \label{eq:V1}
    b^{t+1}(g) = P(g \mid \mathcal{D}) \propto P(a_h^t \mid s^t, g) \cdot b^t(g)
\end{equation}
where $\mathcal{D} = \{(s^1, a_h^1), \ldots (s^t, a_h^t)\}$ is the set of human inputs and $t$ is the current timestep. Notice that \eq{V1} only depends on human actions and not the autonomous assistance --- this prevents the robot from incorrectly learning from itself through a feedback loop. In practice, to evaluate \eq{V1} we need the likelihood function $P(a_h \mid s, g)$, which models how probable it is that the human will choose action $a_h$ given that their goal is $g$ and the robot is in state $s$. Within shared autonomy, this likelihood function commonly follows the Boltzmann-rational model \cite{dragan2013policy}:
\begin{equation} \label{eq:V2}
    P(a_h \mid s, g) = \frac{e^{\,\beta Q_g(s, a_h)}}{\int e^{\,\beta Q_g(s, a)} ~da} \approx \frac{e^{\,\beta Q_g(s, a_h)}}{e^{\,\beta Q_g(s, a_g)}}
\end{equation}
On the far right side we use Laplace's method to approximate the denominator, so that $a_g$ is the optimal action to reach goal $g$. The constant $\beta \geq 0$ is a tunable hyperparameter which determines how sensitive the robot is to each human input. When $\beta \rightarrow 0$ the robot treats the human's actions as random, and when $\beta \rightarrow \infty$ the robot models the human as perfectly rational. We used $\beta = 0.1$ for our user study.

\p{Visual Mapping} Combining Equations~(\ref{eq:V1}) and (\ref{eq:V2}) we have a way to update the robot's belief. Our final step is to map this changing belief into an augmented reality visualization. Although there are an infinite number of valid possibilities, we draw from prior work to identify a natural mapping that requires minimal user interpretation \cite{walker2018communicating, rosen2019communicating, brooks2020visualization}.

Specifically, we map the robot's inferred belief over each goal to the \textit{color} and \textit{transparency} of the corresponding waypoints (see \fig{front}). When the robot is sure that a given $g$ \textit{is not} the human's desired goal, i.e., $b(g) \rightarrow 0$, the trajectory moving to that goal (and the goal itself) are rendered as completely transparent. Conversely, as $b(g) \rightarrow 1$ and the robot becomes more confident that $g$ \textit{is} the human's goal, we linearly interpolate the waypoint color between a dull gray and vibrant orange. This mapping intuitively highlights goals the robot thinks are likely while simultaneously hiding goals that the human does not want.

\subsection{Eliciting Teaching with Active Prompts}

We hypothesize that our augmented reality feedback will reveal the robot's inference while the human is watching the robot (and can see the visual overlays). But what happens when the human looks away to do something else? So far the feedback has been purely \textit{passive} --- providing information only when the human seeks it out. Next we turn to \textit{active} haptic feedback that purposely alerts the human and guides their teaching, even if they are not concentrating on the robot. Wearable haptics is suitable here because it offers direct, attention grabbing signals without encumbering the human.

We break this problem down into two parts: i) determining \textit{when} to provide haptic prompts and ii) selecting \textit{which} haptic prompts will elicit useful human teaching.

\begin{figure*}[t]
	\begin{center}
		\includegraphics[width=2\columnwidth]{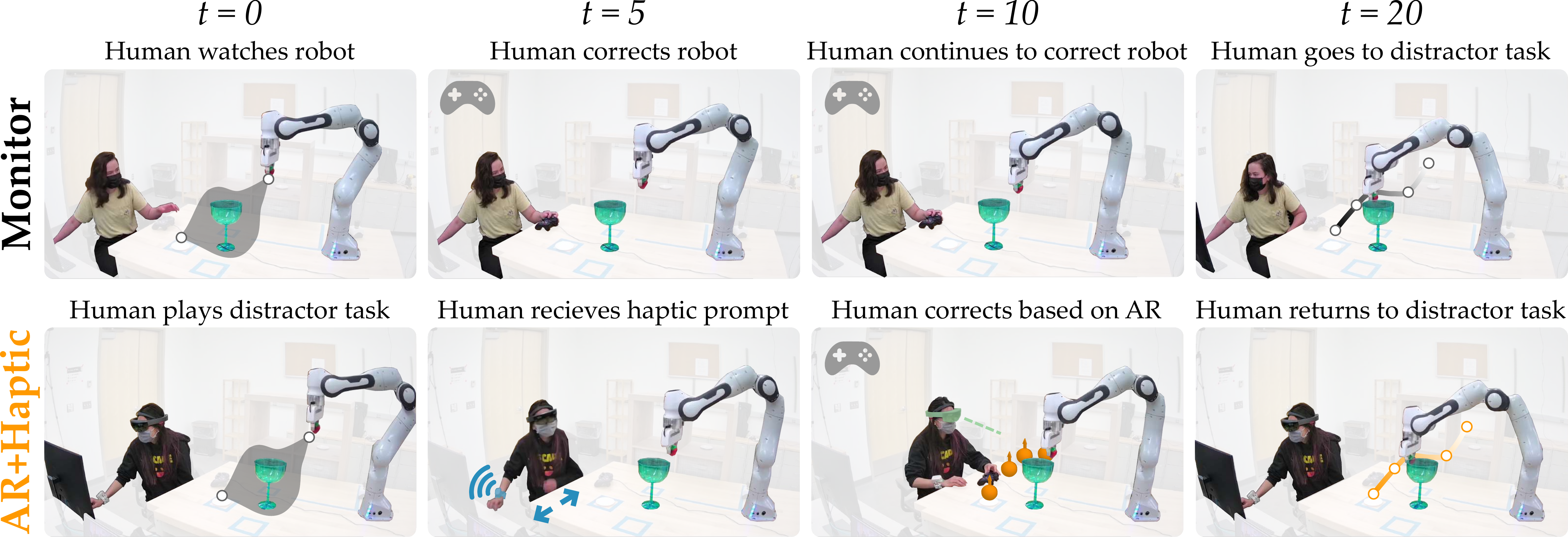}
		\caption{Example robot and human behavior during the \textit{Avoiding} task. The robot is trying to drop a strawberry on the plate, but does not know how to avoid the obstacle. When users receive feedback on the \textbf{GUI} (top row), they are more tentative, and spend extra time watching and correcting the robot. This leads to higher interaction time, lower teaching efficiency, and lower scores on the distractor task. By contrast, humans with \textbf{AR+Haptic} feedback (bottom row) are free to focus on the distractor task. When the robot needs help they get an active haptic notification, and passively observe how the robot learns from their corrections. This multimodal feedback enables the human to efficiently teach the robot without having to constantly monitor its behavior.}
		\label{fig:task}
	\end{center}
    \vspace{-2em}
\end{figure*}

\p{When to Prompt} In our motivating example you are trying to teach the robot to place a coffee cup upright on the lower shelf. At the beginning of the task you notice that the robot is moving towards the shelves --- and since the robot is going in the right direction, you turn away to focus on something else. The robot still needs to figure out whether it should place the cup on the top shelf or the lower shelf; but when is the right time to ask for your feedback? At the start of the task the shelves are far away, and the trajectories to both the top and bottom shelf are very similar. However, as the robot approaches the shelves, the trajectories for the two likely goals diverge, and the robot needs to make a decision (\textit{should I move up or down?}). Intuitively, we want to alert the human at important decision points. 

Drawing from prior research on critical states \cite{huang2018establishing} and inverse reinforcement learning \cite{ramachandran2007bayesian}, we define these decision points as states $s$ where taking the robot's autonomous action $a_r$ will produce a much worse result than acting optimally:
\begin{equation} \label{eq:H1}
    \mathcal{C}(s) = \sum_{g \in \mathcal{G}} b(g) \cdot \big(Q_g(s, a_g) - Q_g(s, a_r) \big)
\end{equation}
Here $\mathcal{C}$ is a measure of how critical the state $s$ is, and if $\mathcal{C}$ exceeds some threshold $\sigma$ then the robot is in a \textit{critical state}. Recall that $a_g$ is the optimal action for goal $g$ (e.g., moving in a straight line from $s$ towards $g$). Our choice of \eq{H1} highlights states where multiple likely goals require very different actions, and the robot needs more information in order to select the correct action.

\p{Which Prompt to Use} We keep track of \eq{H1} during the task, and use the haptic wristband to alert the human when $C(s) > \sigma$. This prompt brings the human's attention back to the robot --- but what feedback should the human teacher provide when they receive an alert? In our running example, the robot has reached the shelves and is unsure whether the cup belongs on the upper or lower shelf. Here the robot would learn the most from an input $a_h$ which moves the robot either \textit{up} or \textit{down}. More generally, we want to encourage the user to teleoperate the robot in the \textit{direction} that best removes uncertainty over the human's goal.

Recall that our haptic device has vibrotactors distributed around the wearer's wrist (see \fig{haptic}). By actuating these vibrotactors in patterns we render direction-based prompts to the human \cite{che2020efficient}. Let $\mathcal{U}$ be the set of prompts we can render: for instance, $u_z \in \mathcal{U}$ alternates between actuating the top and bottom vibrotactors to prompt the user to teleoperate the robot up or down. Given that the robot has reached a critical state, which stimuli $u \in \mathcal{U}$ should we send? Intuitively, we search for the stimulus $u$ which --- if the human responds --- provides the most \textit{information} about the human's unknown goal. Formally, we elicit the direction of human teaching that greedily maximizes the robot's expected information gain:
\begin{equation} \label{eq:H2}
    u^*(s) = \text{arg}\max_{u \in \mathcal{U}} ~ I(g, a_h \mid u, s, b)
\end{equation}
Here $I$ is mutual information and $a_h$ is the human's response to prompt $u$. When solving for $u^*$ we assume that the human will comply with the robot's prompt --- so that if the robot chooses $u_z$, the human will input either $a_h = up$ or $a_h = down$. Expanding \eq{H2}, we obtain \cite{biyik2020learning}:
\begin{multline} \label{eq:H3}
    u^*(s) = \text{arg}\max_{u \in \mathcal{U}} \sum_{a_h \in u}\sum_{g \in \mathcal{G}} b(g)P(a_h \mid u, s, g) \\ \cdot \log_2\frac{P(a_h \mid u, s, g)}{\sum_{g' \in \mathcal{G}}b(g')P(a_h \mid u, s, g)}
\end{multline}
The key to evaluating \eq{H3} is modeling the likelihood function $P(a_h \mid u, s, g)$, which captures how likely it is that the human will respond with action $a_h$ to prompt $u$. In other words, if we signal prompt $u_z$, what is the probability the human applies $a_h=up$ instead of $a_h=down$? Just like before, we leverage the Boltzmann-rational model:
\begin{equation} \label{eq:H4}
    P(a_h \mid u, s, g) = \frac{e^{\beta Q_g(s, a_h)}}{\sum_{a \in u} ~e^{\beta Q_g(s, a)}}
\end{equation}
Comparing \eq{H4} to \eq{V2}, the only difference is our assumption that the human will respond by moving the robot in one of the two discrete directions suggested by $u$. We emphasize that this assumption is purely used to solve for the optimal prompt, and is not imposed in our user study.

\p{Haptic Mapping} Putting it all together, \eq{H1} lets us know \textit{when} to provide haptic prompts, and Equations~(\ref{eq:H3}) and (\ref{eq:H4}) tell us \textit{which} direction-based prompts will elicit the most information. Within the context of our motivating example: as the robot nears the shelves it realizes that it needs more information, and alerts the user by vibrating the haptic wristband on the top and bottom --- encouraging to the human to teleoperate the robot either up or down.

\begin{figure*}[t]
	\begin{center}
		\includegraphics[width=1.8\columnwidth]{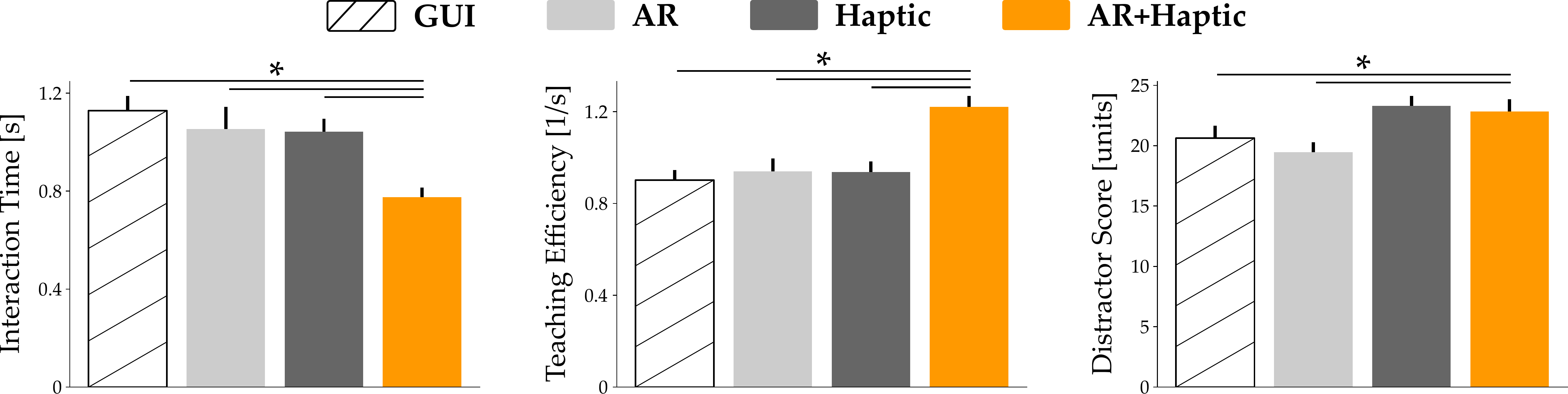}
		\caption{Objective results from our user study. The study included four tasks, and here we report the aggregated results across all tasks. Asterisks denote statistically significant comparisons ($p < .01$) and error bars show standard error. We found that \textbf{AR+Haptic} reduced interaction time and increased teaching efficiency as compared to the alternatives. There was no significant difference in distractor score between \textbf{AR+Haptic} and \textbf{Haptic} ($p = 0.65$).}
		\label{fig:objective}
	\end{center}
    \vspace{-1.75em}
\end{figure*}

\section{User Study}
\label{sec:user}

To test how passive and active multimodal feedback closes the loop during robot inference, we conducted a user study motivated by assistive robotics. Participants interacted with a 2-DoF joystick to teleoperate a 7-DoF robot. The robot shared autonomy with the participant: as the robot inferred the human's goal in real-time during the task, it leveraged different types of feedback to communicate what it had inferred back to the user. We designed experiments where participants had to trade-off between teaching on the robot and focusing on a distractor task.

\begin{figure}[t]
	\begin{center}
		\includegraphics[width=0.8\columnwidth]{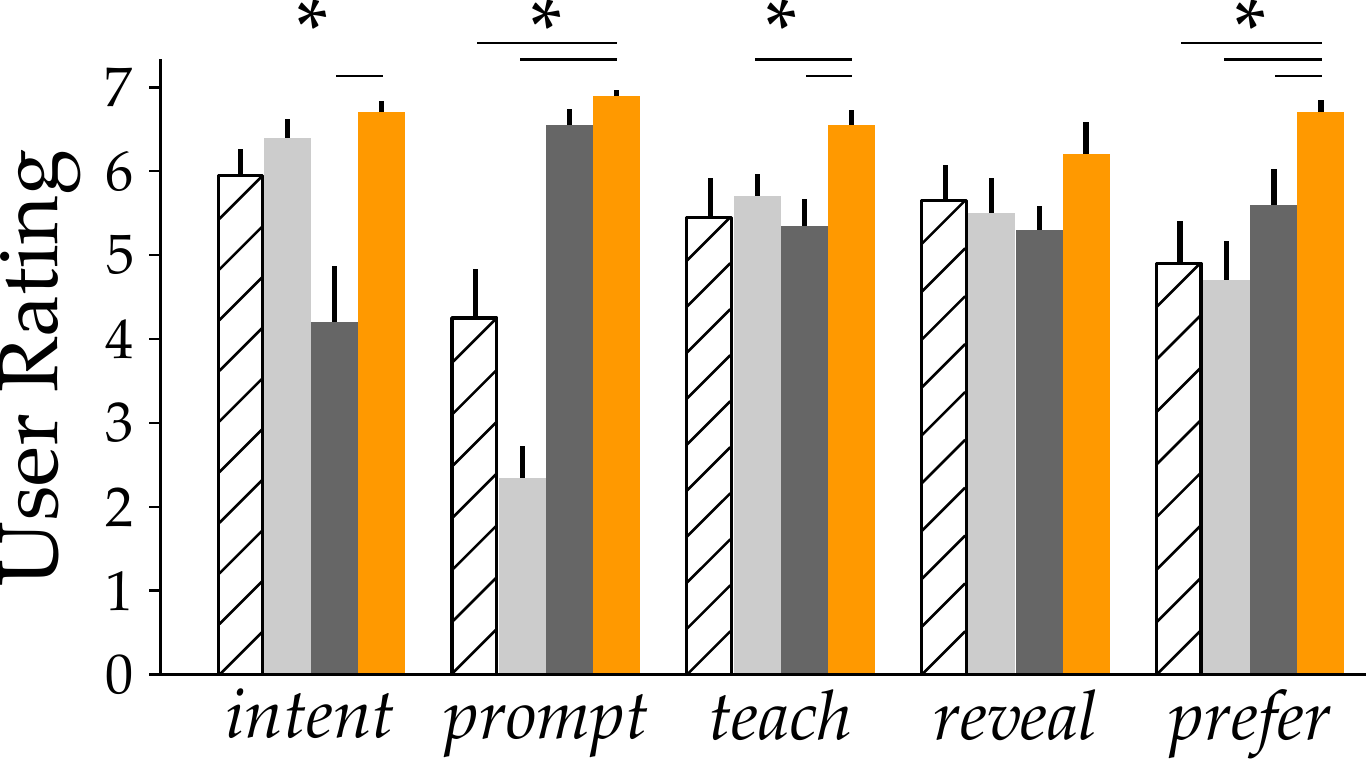}
		\caption{Subjective results from our user study. Colors are consistent with \fig{objective}, and higher ratings indicate agreement (i.e., conveys intent, prompts for help). Overall, participants preferred teaching the robot with multimodal \textbf{AR+Haptic} feedback. We note that some users reported the \textbf{AR} device to be uncomfortable, which may skew their perception of this condition.}
		\label{fig:subjective}
	\end{center}
    \vspace{-2em}
\end{figure}

\p{Independent Variables} We compared four different feedback conditions: 
\begin{itemize}
    \item A graphical user interface (\textbf{GUI}) on the computer screen
    \item Passive augmented reality feedback (\textbf{AR})
    \item Active haptic feedback (\textbf{Haptic})
    \item Augmented reality + haptic feedback (\textbf{AR+Haptic})
\end{itemize}
Within the \textbf{GUI} baseline we displayed a graphical user interface on the same computer screen as the distractor task. This \textbf{GUI} provided both active and passive feedback: we displayed the robot's belief over each goal, and printed a text alert when the robot reached a critical state. The \textbf{AR} and \textbf{Haptic} baselines are suitable for purely passive or active feedback. In all cases where we used \textbf{Haptic} to provide active cues the participants responded by teaching the robot. Similarly, participants with \textbf{AR} feedback were confident that they understood the robot's intent (see \fig{subjective}). Finally, the \textbf{AR+Haptic} condition is our proposed approach from Section~\ref{sec:method} that combines both passive and active feedback.

\p{Experimental Setup} Participants completed four tasks with every feedback condition. In the \textit{Placing} task (see \fig{front}) participants taught the robot to place a coffee cup on the shelf. There were $8$ candidate goals the robot considered: the front or back of the top shelf, the front or back of the bottom shelf, and putting the cup down either upright or on its side. In the \textit{Avoiding} task (see \fig{task}) users adjusted the robot's motion to avoid an obstacle. The robot initially moved directly towards this obstacle and was unsure if it needed to avoid it, and if so, which way it should go around. Here the robot had $4$ discrete candidate goals: moving through, to the left, to the right, or above the obstacle. Both the \textit{Sorting} task and the \textit{Forgetting} task started in the same way with the robot carrying a bottle of glue towards the shelves. The robot had $4$ candidate goals corresponding to different locations on the shelf. In the \textit{Sorting} task the user taught the robot to sort the glue on the lower shelf. By contrast, half-way through the \textit{Forgetting} task the robot automatically became confused about the human’s goal (i.e., it forgot what it had learned), so that it required additional teaching.


During all four tasks users divided their attention between teaching the robot and playing a distractor game. Within this game participants tried to track a moving target with their mouse, and accrued points when their mouse was inside the target. Because the game was played on a screen orthogonal to the robot, users could not focus on both the game and robot at the same time. The set of haptic cues $\mathcal{U}$ included three options: i) alternating the top and bottom vibrotactors, ii) alternating the left and right vibrotactors, and iii) actuating all vibrotactors in a circular pattern.

\p{Participants and Procedure} We recruited ten participants from the Virginia Tech community to take part in our study (4 female, average age $24.3 \pm 3.0$ years). All subjects provided informed written consent prior to the experiment. Half of the participants had previous experience interacting with a robot arm, and two participants had used augmented reality before. Participants teleoperated the robot for up to five minutes to familiarize themselves with each feedback method. We used a within-subjects study design: every participant interacted with all four feedback conditions. To mitigate the confounding effect of participants improving over time, we counterbalanced the order of the feedback conditions.

\p{Dependent Measures -- Objective} We recorded the amount of time users spent teaching the robot with the joystick (\textit{Interaction Time}), as well as the total score users achieved on the distractor game (\textit{Distractor Score}). We also measured the user's \textit{Teaching Efficiency}. To compute this we recorded $b(g)$, the robot's learned belief in the human's desired goal at the end of the task. Recall that $b(g)$ is a scalar, where $b(g) \rightarrow 1$ indicates that the robot correctly inferred the human’s goal. Following \cite{sena2020quantifying}, we then divided this scalar by the user's interaction time (i.e., the amount of time they teleoperated the robot). Intuitively, a high \textit{Teaching Efficiency} indicates that the human accurately taught the robot their desired goal through a smaller number of well-timed corrections.

\p{Dependent Measures -- Subjective} Participants filled out a 7-point Likert scale survey after each method. Questions were grouped into five multi-item scales (see \fig{subjective}): did the user understand the robot's \textit{intent}, did the user know when the robot needed more teaching (\textit{prompt}), did the robot make it clear what type of \textit{teaching} to provide, did the robot \textit{reveal} what it knew and did not know, and did the user \textit{prefer} this current feedback method to the alternatives.

\p{Hypotheses} We had three hypotheses in this user study:
\begin{displayquote}
    \textbf{H1.} \emph{Users with passive visual feedback (\textbf{AR}) will minimize interaction time.}
\end{displayquote}
\begin{displayquote}
    \textbf{H2.} \emph{Users with active haptic feedback (\textbf{Haptic}) will maximize score on the distractor task.}
\end{displayquote}
\begin{displayquote}
    \textbf{H3.} \emph{Users with multimodal feedback (\textbf{AR+Haptic}) will maximize teaching efficiency.}
\end{displayquote}

\p{Results -- Objective} An example interaction is illustrated in \fig{task}, and our aggregate results across all tasks are plotted in \fig{objective}. When interpreting these results, remember that we want to minimize interaction time while maximizing teaching efficiency and distractor score.

Because \textbf{AR} displays what the robot has inferred, we anticipated that with \textbf{AR} participants would know when the robot understood their goal, and would only interact as much as needed. More specifically, we anticipated that \textbf{AR} would minimize interaction time. This turned out not to be true: only in the \textit{Avoiding} task did \textbf{AR} reduce the interaction time. When looking at the aggregated results, post hoc analysis shows no significant differences in interaction time between \textbf{AR} and \textbf{GUI} ($p=.55$) or \textbf{AR} and \textbf{Haptic} ($p=.93$).

Unlike passive \textbf{AR}, active \textbf{Haptic} feedback grabs the human's attention when teaching is required. Since this method provides alerts, participants can focus on the distractor game until they receive a prompt. To test \textbf{H2}, we compared the distractor score with and without \textbf{Haptic}. We found that in either the \textbf{Haptic} or \textbf{AR+Haptic} condition participants scored statistically significantly higher than with \textbf{GUI} or \textbf{AR} ($p < .01$). This suggests that active feedback reduces the amount of time users spent monitoring the robot.

Thus far we have looked at the results of \textbf{AR} and \textbf{Haptic} individually; what happens when we combine these into multimodal \textbf{AR+Haptic} feedback? Inspecting \fig{objective}, participants using \textbf{AR+Haptic} had lower interaction time and higher teaching efficiency than the alternatives ($p < .01$). These results support \textbf{H3}. We conclude that the synthesis of passive and active feedback was more effective for communicating robot inference than either method in isolation.

\p{Results -- Subjective} The results from our 7-point Likert scale survey are listed in \fig{subjective}. We first tested the reliability of all five scales using Cronbach’s alpha, and found every scale to be reliable ($\alpha$ > 0.7). Accordingly, we grouped the participants’ responses within each scale into a single combined score, and ran a one-way repeated measures ANOVA. Post-hoc tests revealed that users preferred \textbf{AR+Haptic} to the alternatives ($p < .05$). One user commented that \textit{``the combination of tactile and visual feedback made this method the easiest to control the robot.''}

\p{Limitations} During our user study the \textbf{AR} was implemented on a Microsoft HoloLens. Several participants indicated that they disliked wearing this device for prolonged periods: ``\textit{the HoloLens is uncomfortable to wear.''} We recognize that user comfort may be a confounding factor that negatively affected \textbf{AR} results. In addition, our results indicate that providing passive and active feedback on the \textbf{GUI} was roughly on par with passive feedback in \textbf{AR}. This suggests that a \textbf{GUI+Haptic} system (with passive visual feedback and active haptic prompts) could potentially achieve similar performance as \textbf{AR+Haptic}. However, here the GUI must be located in a position where human teachers can readily view it while interacting with the robot arm. Our future work will explore intuitive \textbf{GUI} design for \textbf{GUI+Haptic} systems.
\section{Conclusion}

From the human's perspective robot inference is often a black box. Without any feedback, humans have no way of knowing what their robot has inferred or whether it is confused. Accordingly, we developed a multimodal augmented reality and haptic interface that combines \textit{passive} and \textit{active} feedback. We derived a decision rule for mapping robot inference to each type of feedback during shared autonomy tasks. Our user study demonstrates that this passive and active approach takes a step towards bringing the human into the loop during robot inference.


\balance
\bibliographystyle{IEEEtran}
\bibliography{IEEEabrv,bibtex}

\begin{thebibliography}{10}
\providecommand{\url}[1]{#1}
\csname url@rmstyle\endcsname
\providecommand{\newblock}{\relax}
\providecommand{\bibinfo}[2]{#2}
\providecommand\BIBentrySTDinterwordspacing{\spaceskip=0pt\relax}
\providecommand\BIBentryALTinterwordstretchfactor{4}
\providecommand\BIBentryALTinterwordspacing{\spaceskip=\fontdimen2\font plus
\BIBentryALTinterwordstretchfactor\fontdimen3\font minus
  \fontdimen4\font\relax}
\providecommand\BIBforeignlanguage[2]{{%
\expandafter\ifx\csname l@#1\endcsname\relax
\typeout{** WARNING: IEEEtran.bst: No hyphenation pattern has been}%
\typeout{** loaded for the language `#1'. Using the pattern for}%
\typeout{** the default language instead.}%
\else
\language=\csname l@#1\endcsname
\fi
#2}}

\bibitem{hellstrom2018understandable}
T.~Hellstr{\"o}m and S.~Bensch, ``Understandable robots -- {W}hat, why, and
  how,'' \emph{Journal of Behavioral Robotics}, vol.~9, pp. 110--123, 2018.

\bibitem{dragan2013legibility}
A.~D. Dragan, K.~C. Lee, and S.~S. Srinivasa, ``Legibility and predictability
  of robot motion,'' in \emph{ACM/IEEE Int. Conf. on Human-Robot Interaction},
  2013, pp. 301--308.

\bibitem{tellex2014asking}
S.~Tellex, R.~Knepper, A.~Li, D.~Rus, and N.~Roy, ``Asking for help using
  inverse semantics,'' in \emph{Robotics: Science and Systems}, 2014.

\bibitem{admoni2016predicting}
H.~Admoni and S.~Srinivasa, ``Predicting user intent through eye gaze for
  shared autonomy,'' in \emph{AAAI}, 2016.

\bibitem{cha2018survey}
E.~Cha, Y.~Kim, T.~Fong, and M.~J. Mataric, ``A survey of nonverbal signaling
  methods for non-humanoid robots,'' \emph{Foundations and Trends in Robotics},
  vol.~6, no.~4, pp. 211--323, 2018.

\bibitem{jain2019probabilistic}
S.~Jain and B.~Argall, ``Probabilistic human intent recognition for shared
  autonomy in assistive robotics,'' \emph{ACM Transactions on Human-Robot
  Interaction}, vol.~9, no.~1, pp. 1--23, 2019.

\bibitem{javdani2018shared}
S.~Javdani, H.~Admoni, S.~Pellegrinelli, S.~S. Srinivasa, and J.~A. Bagnell,
  ``Shared autonomy via hindsight optimization for teleoperation and teaming,''
  \emph{The Int. Journal of Robotics Research}, vol.~37, no.~7, pp. 717--742,
  2018.

\bibitem{dragan2013policy}
A.~D. Dragan and S.~S. Srinivasa, ``A policy-blending formalism for shared
  control,'' \emph{The Int. Journal of Robotics Research}, vol.~32, no.~7, pp.
  790--805, 2013.

\bibitem{szafir2014communication}
D.~Szafir, B.~Mutlu, and T.~Fong, ``Communication of intent in assistive free
  flyers,'' in \emph{ACM/IEEE Int. Conf. on Human-Robot Interaction}, 2014, pp.
  358--365.

\bibitem{knepper2017implicit}
R.~A. Knepper, C.~I. Mavrogiannis, J.~Proft, and C.~Liang, ``Implicit
  communication in a joint action,'' in \emph{ACM/IEEE Int. Conf. on
  Human-Robot Interaction}, 2017, pp. 283--292.

\bibitem{gielniak2011generating}
M.~J. Gielniak and A.~L. Thomaz, ``Generating anticipation in robot motion,''
  in \emph{IEEE Int. Symp. on Robot and Human Interactive Communication}, 2011,
  pp. 449--454.

\bibitem{andersen2016projecting}
R.~S. Andersen, O.~Madsen, T.~B. Moeslund, and H.~B. Amor, ``Projecting robot
  intentions into human environments,'' in \emph{IEEE Int. Symp. on Robot and
  Human Interactive Communication}, 2016.

\bibitem{baraka2016enhancing}
K.~Baraka, S.~Rosenthal, and M.~Veloso, ``Enhancing human understanding of a
  mobile robot's state and actions using expressive lights,'' in \emph{IEEE
  Int. Symp. on Robot and Human Interactive Communication}, 2016, pp. 652--657.

\bibitem{weng2019robot}
T.~Weng, L.~Perlmutter, S.~Nikolaidis, S.~Srinivasa, and M.~Cakmak, ``Robot
  object referencing through legible situated projections,'' in \emph{Int.
  Conf. on Robotics and Automation}, 2019, pp. 8004--8010.

\bibitem{newbury2021visualizing}
R.~Newbury, A.~Cosgun, T.~Crowley-Davis, W.~P. Chan, T.~Drummond, and E.~Croft,
  ``Visualizing robot intent for object handovers with augmented reality,''
  \emph{arXiv preprint arXiv:2103.04055}, 2021.

\bibitem{hoang2021virtual}
K.~C. Hoang, W.~P. Chan, S.~Lay, A.~Cosgun, and E.~Croft, ``Virtual barriers in
  augmented reality for safe and effective human-robot cooperation in
  manufacturing,'' \emph{arXiv preprint arXiv:2104.05211}, 2021.

\bibitem{walker2018communicating}
M.~Walker, H.~Hedayati, J.~Lee, and D.~Szafir, ``Communicating robot motion
  intent with augmented reality,'' in \emph{ACM/IEEE Int. Conf. on Human-Robot
  Interaction}, 2018, pp. 316--324.

\bibitem{rosen2019communicating}
E.~Rosen, D.~Whitney, E.~Phillips, G.~Chien, J.~Tompkin, G.~Konidaris, and
  S.~Tellex, ``Communicating and controlling robot arm motion intent through
  mixed-reality head-mounted displays,'' \emph{The Int. Journal of Robotics
  Research}, vol.~38, no. 12-13, pp. 1513--1526, 2019.

\bibitem{muhammad2019creating}
F.~Muhammad, A.~Hassan, A.~Cleaver, and J.~Sinapov, ``Creating a shared reality
  with robots,'' in \emph{ACM/IEEE Int. Conf. on Human-Robot Interaction},
  2019, pp. 614--615.

\bibitem{chandan2021arroch}
K.~Chandan, V.~Kudalkar, X.~Li, and S.~Zhang, ``{ARROCH: Augmented reality for
  robots collaborating with a human},'' in \emph{Int. Conf. on Robotics and
  Automation}, 2021.

\bibitem{pezent2019tasbi}
E.~Pezent, A.~Israr, M.~Samad, S.~Robinson, P.~Agarwal, H.~Benko, and
  N.~Colonnese, ``Tasbi: {M}ultisensory squeeze and vibrotactile wrist haptics
  for augmented and virtual reality,'' in \emph{IEEE World Haptics Conf.},
  2019.

\bibitem{aggravi2018design}
M.~Aggravi, F.~Paus{\'e}, P.~R. Giordano, and C.~Pacchierotti, ``Design and
  evaluation of a wearable haptic device for skin stretch, pressure, and
  vibrotactile stimuli,'' \emph{IEEE Robotics and Automation Letters}, vol.~3,
  no.~3, pp. 2166--2173, 2018.

\bibitem{arrieta2020explainable}
A.~B. Arrieta, N.~D{\'\i}az-Rodr{\'\i}guez, J.~Del~Ser, A.~Bennetot, S.~Tabik,
  A.~Barbado, S.~Garc{\'\i}a, S.~Gil-L{\'o}pez, D.~Molina, and R.~Benjamins,
  ``Explainable artificial intelligence ({XAI}): Concepts, taxonomies,
  opportunities and challenges toward responsible ai,'' \emph{Information
  Fusion}, vol.~58, pp. 82--115, 2020.

\bibitem{huang2019enabling}
S.~H. Huang, D.~Held, P.~Abbeel, and A.~D. Dragan, ``Enabling robots to
  communicate their objectives,'' \emph{Autonomous Robots}, vol.~43, no.~2, pp.
  309--326, 2019.

\bibitem{brooks2020visualization}
C.~Brooks and D.~Szafir, ``Visualization of intended assistance for acceptance
  of shared control,'' in \emph{IEEE/RSJ Int. Conf. on Intelligent Robots and
  Systems}, 2020.

\bibitem{zolotas2018head}
M.~Zolotas, J.~Elsdon, and Y.~Demiris, ``Head-mounted augmented reality for
  explainable robotic wheelchair assistance,'' in \emph{IEEE/RSJ Int. Conf. on
  Intelligent Robots and Systems}, 2018.

\bibitem{huang2018establishing}
S.~H. Huang, K.~Bhatia, P.~Abbeel, and A.~D. Dragan, ``Establishing appropriate
  trust via critical states,'' in \emph{IEEE/RSJ Int. Conf. on Intelligent
  Robots and Systems}, 2018, pp. 3929--3936.

\bibitem{ramachandran2007bayesian}
D.~Ramachandran and E.~Amir, ``Bayesian inverse reinforcement learning,'' in
  \emph{IJCAI}, 2007, pp. 2586--2591.

\bibitem{che2020efficient}
Y.~Che, A.~M. Okamura, and D.~Sadigh, ``Efficient and trustworthy social
  navigation via explicit and implicit robot--human communication,'' \emph{IEEE
  Trans. on Robotics}, vol.~36, no.~3, pp. 692--707, 2020.

\bibitem{biyik2020learning}
E.~B{\i}y{\i}k, D.~P. Losey, M.~Palan, N.~C. Landolfi, G.~Shevchuk, and
  D.~Sadigh, ``Learning reward functions from diverse sources of human
  feedback: {O}ptimally integrating demonstrations and preferences,'' \emph{The
  Int. Journal of Robotics Research}, 2021.

\bibitem{sena2020quantifying}
A.~Sena and M.~Howard, ``Quantifying teaching behavior in robot learning from
  demonstration,'' \emph{The International Journal of Robotics Research},
  vol.~39, no.~1, pp. 54--72, 2020.

\end{thebibliography}

\end{document}